\documentclass[letterpaper, 10 pt, journal, twoside]{IEEEtran}
\usepackage{cite}

\usepackage{subfig}
\ifCLASSINFOpdf
   \usepackage[pdftex]{graphicx}
\else
\fi
\usepackage{amsmath}
\usepackage[linesnumbered,ruled,vlined]{algorithm2e} 

\usepackage{array}
\usepackage{fixltx2e}

\usepackage{stfloats}
\ifCLASSOPTIONcaptionsoff
 \usepackage[nomarkers]{endfloat}
\let\MYoriglatexcaption\caption
\renewcommand{\caption}[2][\relax]{\MYoriglatexcaption[#2]{#2}}
\fi
\usepackage{url}
\usepackage[colorlinks]{hyperref}

\begin{document}
%
\title{Elephants Don't Pack Groceries$^{3}$: Robot Task Planning for Low Entropy Belief States}

\author{Alphonsus Adu-Bredu$^{1}$ \hspace{0.5cm} Zhen Zeng$^{2}$ \hspace{0.5cm} Neha Pusalkar$^{1}$ \hspace{0.5cm} Odest Chadwicke Jenkins$^{1}$%
\thanks{Manuscript received: April, 29, 2021; Revised July, 31, 2021; Accepted September, 1, 2021.}
\thanks{This paper was recommended for publication by
Editor Hanna Kurniawati upon evaluation of the Associate Editor and Reviewers’
comments.} 
\thanks{$^{1}$A.~Adu-Bredu, N.~Pusalkar and O.C.~Jenkins are with the Robotics Institute and Department of Electrical Engineering and Computer Science, University of Michigan, Ann Arbor, MI, USA  
        {\tt\small [adubredu|nehagp|ocj]@umich.edu}}%
\thanks{$^{2}$Z.~Zeng is with J.P. Morgan AI Research. {\tt\small zhen.zeng@jpmchase.com}. This work was completed independently from J.P. Morgan AI Research.%
}
\thanks{$^{3}$ A word-play on Rodney Brooks's seminal work, \textit{Elephants don't play chess}, that inspired the Embodied Intelligence perspective of this work}%
\thanks{Digital Object Identifier (DOI): see top of this page.}
} 
%
%

\markboth{IEEE Robotics and Automation Letters. Preprint Version. Accepted September, 2021}
{Adu-Bredu \MakeLowercase{\textit{et al.}}: lesample} 

%



\maketitle

\begin{abstract}
Recent advances in computational perception have significantly improved the ability of autonomous robots to perform state estimation with low entropy. 
Such advances motivate a reconsideration of robot decision-making under uncertainty. 
Current approaches to solving sequential decision-making problems model states as inhabiting the extremes of the perceptual entropy spectrum.
As such, these methods are either incapable of overcoming perceptual errors or asymptotically inefficient in solving problems with low perceptual entropy.
With low entropy perception in mind, we aim to explore a happier medium that balances computational efficiency with the forms of uncertainty we now observe from modern robot perception.  We propose an approach for efficient task planning for goal-directed robot reasoning. 
Our approach combines belief space representation  with the fast, goal-directed features of classical planning to efficiently plan for low entropy goal-directed reasoning tasks.
We compare our approach with current classical planning and belief space planning approaches by solving low entropy goal-directed grocery packing tasks in simulation. 
Our approach outperforms these approaches in  planning time, execution time, and task success rate in our simulation experiments. We also demonstrate our approach on a real world grocery packing task with physical robot. A video summary of this paper can be found at this url: {\url{https://youtu.be/im6tve9-9A0}}
\end{abstract}


%
\IEEEpeerreviewmaketitle

\section{Introduction}
%
%
%
%
\IEEEPARstart{S}{equential} decision-making problems have often been modelled as either fully-observable or partially observable. Fully observable models have no entropy in states and actions whilst partially observable models have high entropy in states and actions.
With these models have come classical planning approaches~\cite{a6,a11,b10,a12} for solving zero entropy problems and belief space planning approaches~\cite{pomcp,beliefpddlstream,a2} for solving high entropy problems. 
Visualizing entropy as a spectrum, classical planning approaches plan with models on one extreme end of the entropy spectrum whilst belief space planning approaches plan with models on the other extreme.
Recent advances in robot perception systems, both in terms of inexpensive hardware~\cite{realsense} and fast and efficient algorithms~\cite{e1,e2,e3,e4,e5} have significantly  reduced the state estimation entropy when used for robot manipulation.
When a robot is equipped with such a low entropy perception system, the robot’s sequential decision-making problem does not fall at either extremes of the entropy spectrum. 
The problem falls in an intermediate region on the spectrum where neither family of approaches are equipped to exploit the low entropy nature of the problem to solve it efficiently.

\begin{figure}[t]
    \centering
    \includegraphics[width=0.48\textwidth]{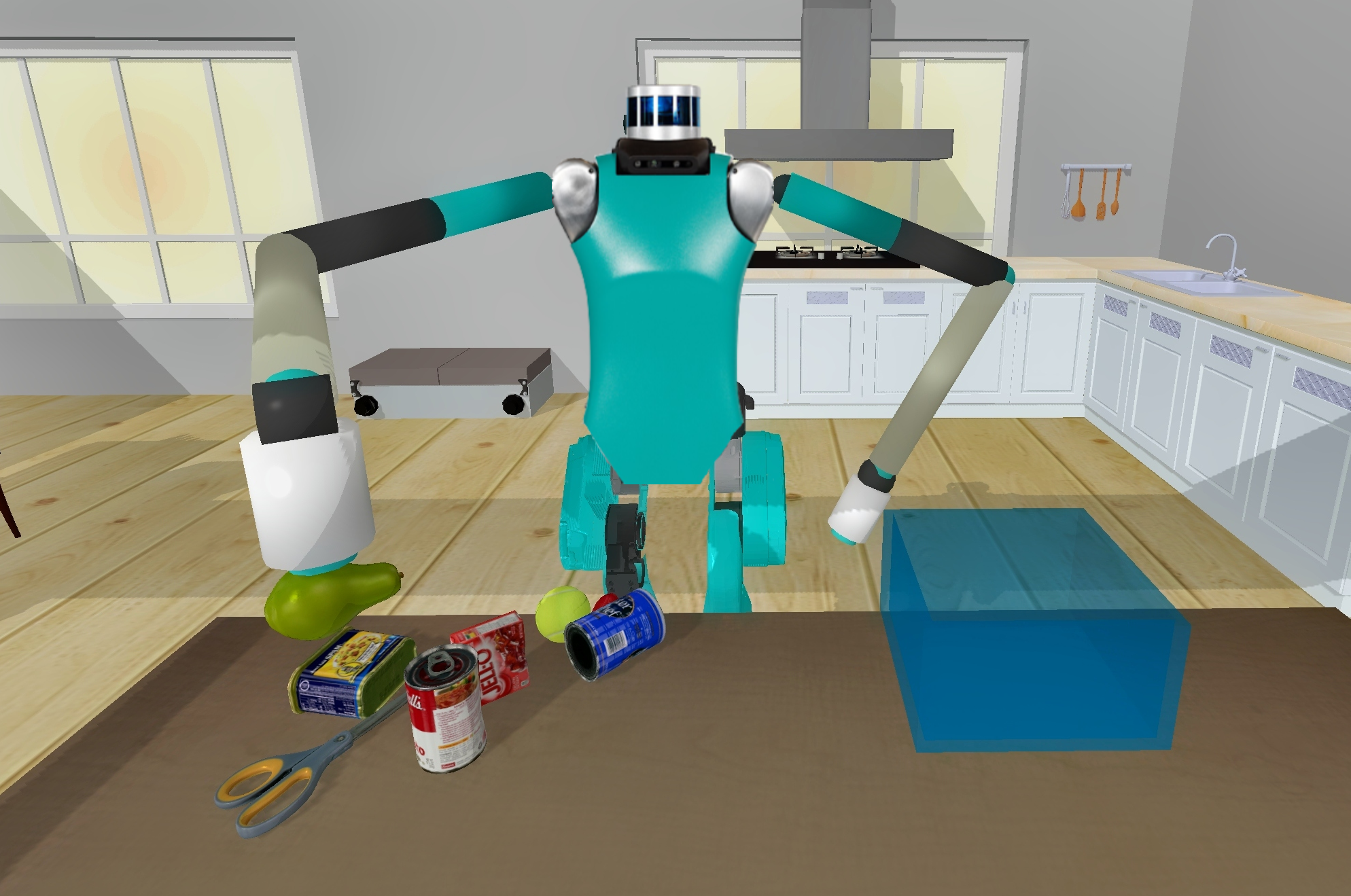}
    \caption{
    A sequential grocery packing task using LESAMPLE. The simulated Digit robot equipped with suction grippers is able to efficiently pack the groceries under low perceptual entropy to satisfy given goal constraints. }
    \label{fig:1}
\end{figure}

Classical planning approaches do not account for uncertainty so they often fail to generate feasible plans in uncertain domains. Some belief space planning approaches attempt to exactly solve for the optimal policy that maps belief states to actions \cite{vi}. Since solving for the optimal policy exactly  becomes intractable for realistic problems, other belief space planning approaches approximate the belief space through sampling \cite{pomcp, despot}. Although these approximate methods are tractable, they tend to be inefficient for low entropy state spaces.


Results from early work in Embodied Intelligence by Brooks et. al. \cite{elephants, aim} demonstrate that, methods that plan and act on loose models of the world and rely on sensor feedback to adjust their behavior are often more efficient and practical than their counterparts that perform explicit modelling of all possibilities before taking an action. These results are also echoed in more recent work \cite{b6, b7} that show replanning approaches to be more efficient in domains with stochastic action effects than probabilistic planning approaches. Inspired by these results, we hypothesize that for a state space with low perceptual entropy, a simple replanning approach that samples from the belief space and plans using this sample will be more efficient than belief space planning in solving the task at hand.

In light of this, we propose a decoupled approach to goal-directed robotic manipulation that builds on the respective strengths of classical and belief space planning.  As motivated by Sui et al.~\cite{e1}, task planning can be performed on state estimates from perceived belief distributions, and updated when the perceptual probability mass shifts to a different state estimate. 
Building on this idea and recent work in replanning algorithms~\cite{b6}, we propose \textbf{Low Entropy Sampling planner} (\textbf{LESAMPLE}) as a simple and efficient online task planning algorithm for solving problems with low entropy in state estimation and deterministic action effects.  
 

The concept of replanning with estimates from the belief space is not novel and has been employed in works like Yoon et. al. \cite{b6}. This paper does not claim to propose an entirely new algorithm. We instead aim to demonstrate the efficiency benefits a simple replanning approach could have over belief space planning methods, when used to solve low entropy planning problems.

We benchmark LESAMPLE against current classical planning and belief space planning approaches by solving low entropy goal-directed grocery packing tasks in simulation as shown in Figure \ref{fig:1}. 
LESAMPLE outperforms these approaches with respect to  planning time, execution time, and task success rate in our simulation experiments.


\begin{figure}[t]
    \centering
    \includegraphics[width=0.48\textwidth]{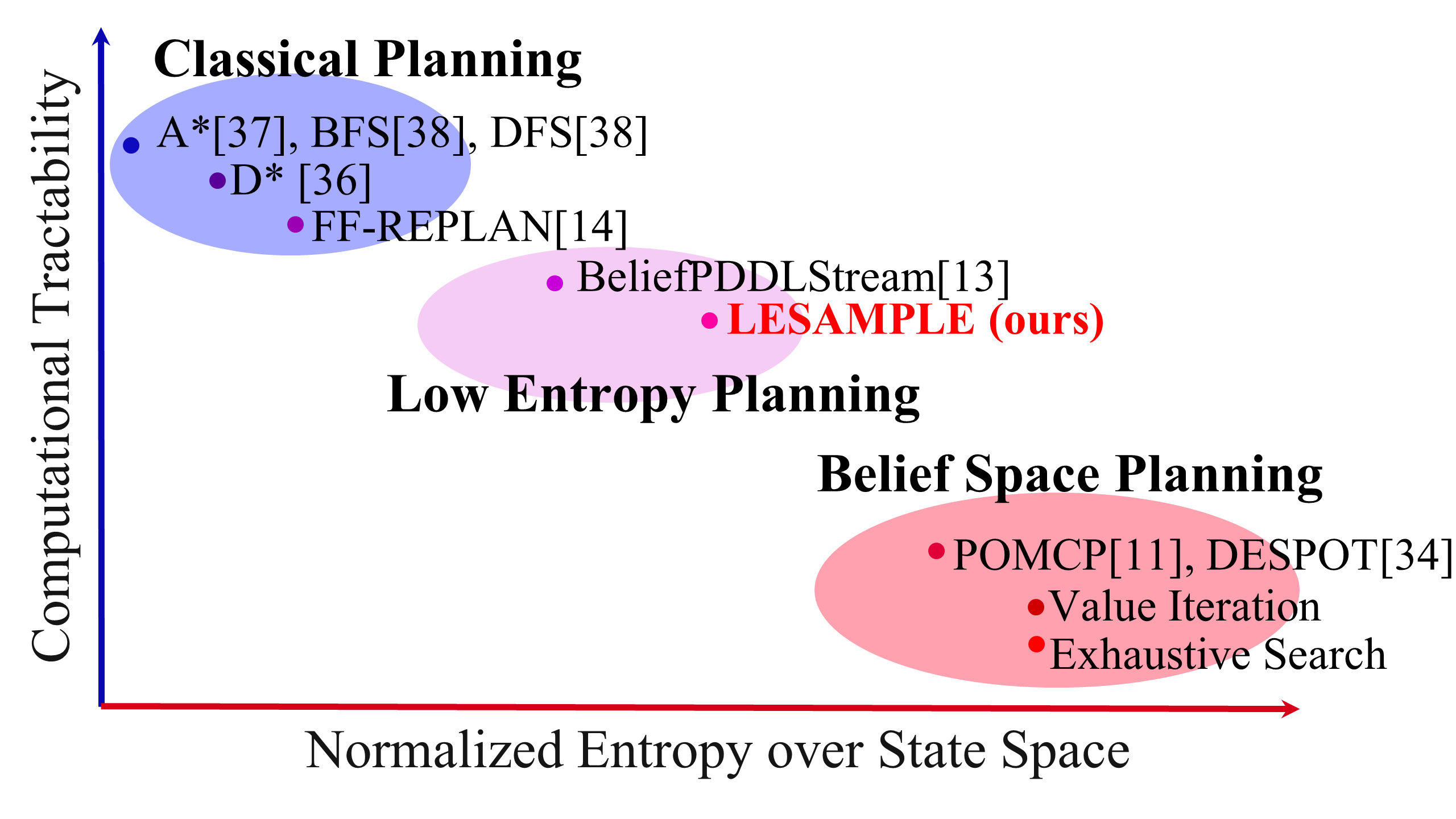}
    \caption{ \small An illustration comparing the relative the computational tractability of planning algorithms and the level of perceptual entropy in problems they are designed to solve. Classical planning algorithms are computationally tractable (fast) and are designed to solve problems with no entropy in their state space. Belief space planning algorithms are generally slow on reasonably complex problems and are designed to solve problems with high perceptual entropy in their state space. Our algorithm, LESAMPLE, is designed to efficiently solve problems with low perceptual entropy. }%
    \label{fig:2}%
\end{figure}
%
\section{RELATED WORK}
\subsection{Classical Planning}
Classical planning approaches ~\cite{astar, clrs, dstar} are used to solve fully-observable and deterministic problems.
These approaches are fast and usually come with convergence and optimality guarantees, making them convenient to use on suitable problems.
They however do not account for entropy when planning so they often generate infeasible plans in uncertain domains. 

To solve problems of a sequential nature such as grocery packing, the robot has to be able to reason over both symbolic states of the world  as well as continuous states. This family of problems is known as Task and Motion Planning Problems~\cite{a1,a2,a4,a6,a8,a10}. Works such as Kaelbling et. al.~\cite{a11}, Srivastava et. al.~\cite{a1} and  Garrett et. al.~\cite{a10} have focused on ways to interleave the symbolic planning involved in task planning with the continuous-space planning involved in motion planning in order for the robot to generate feasible plans and actions. In our proposed LESAMPLE method, we first generate a symbolic plan and later use continuous parameter sampling and sampling based motion planning \cite{birrt} to generate continuous trajectories for performing the task at hand. 

\subsection{Belief Space Planning}
Sequential decision-making problems with high entropy in state estimation and action effects are often modelled as Partially Observable Markov Decision Processes (POMDP)~\cite{b1}.
The states in a POMDP are probability distributions called belief states. 
To solve a POMDP is to find an optimal policy that maps belief states to actions.
However solving POMDPs exactly is intractable due to the curse of dimensionality ~\cite{b1} and the curse of history~\cite{b4}.
As such, belief space planning methods such as POMCP~\cite{pomcp} and DESPOT~\cite{despot} are used to solve POMDPs approximately. 
POMCP~\cite{pomcp} uses Monte Carlo Sampling to sample the belief state  and belief transitions and uses Monte Carlo Tree Search~\cite{b8} to search for an optimal policy. DESPOT~\cite{despot} improves upon POMCP's worst case behavior by sampling a small number of scenarios and performing search over a determinized sparse partially observable tree.
Our proposed approach represents each state as a set of hypotheses which are weighted based on the robot's observation. We use weighted sampling~\cite{wsample} to sample from the weighted hypotheses to get a reliable estimate of the states. We evaluate LESAMPLE against POMCP and DESPOT in our experiments.

\subsection{Integrating Belief Space Representation with Classical Planning}
FF-Replan~\cite{b6}, a classical planning approach, attempts to solve problems with no entropy in state estimation but high entropy in action effects by constantly replanning. 
FF-Replan determinizes the action effects through choosing the effect with the highest confidence. It then applies Fast-Forward~\cite{a12} to plan in the determinized domain and re-plans whenever there is an inconsistency caused by the disregard of the entropy in the domain.
This algorithm is shown to work quite well for certain problems and terribly for others depending on how well the determinization reflects the true action effects of the domain. 
Other approaches such as BeliefPDDLStream~\cite{beliefpddlstream} use particles to represent the belief space and update the particles after each observation using a particle filter and replans using this estimate of the belief space. 
Our proposed approach represents each state as a set of weighted hypotheses and use weighted sampling~\cite{wsample} to sample from the weighted hypotheses to get a reliable estimate of the states. 
We then employ symbolic planning to plan in the sampled states and reweight, resample and replan when needed. Figure \ref{fig:2} shows a graphical comparison of LESAMPLE with other classical and belief space planning methods. We also evaluate LESAMPLE against FF-Replan and BeliefPDDLStream in our experiments.

\subsection{Embodied Intelligence}
Early research in Embodied Intelligence \cite{aim, elephants} demonstrated the efficiency of methods  that  plan and  act  on  loose  models  of  the  world  and  rely  on  sensor feedback  to  adjust  their  behavior. These approaches worked best for domains where execution errors were reversible. More recent work in planning under uncertainty \cite{b6, b7} have also produced results that show the efficiency of replanning approaches over deliberate probabilistic planning methods in partially observable domains. These early work also gave rise to approaches \cite{actsense1, actsense2,actsense4} that explicitly perform information gathering actions and decide the next best action based on the obtained sensory information. LESAMPLE updates its belief after every action taken and replans whenever the updated belief doesn't match the predicted effect of the executed action.

\subsection{Bin Packing}
A vast body of work has addressed the robot bin packing problem~\cite{c2,c8,c9}. Amongst these, Wang and Hauser ~\cite{c2}, Weng et. al.~\cite{c9} consider the problem from a geometric perspective and try to find the optimal packing arrangement of objects such that they use up a minimum number of bins and a minimum amount of space.
Wang and Hauser ~\cite{c8} goes further to optimize for space-efficient packing arrangements that result in stable object piles. With the grocery packing task considered in this work, our proposed approach, LESAMPLE, does not explicitly optimize for efficient space usage when packing. We mainly just sample free placement poses in the destination bin that are large enough for the item in hand to be placed at.  


\section{PROBLEM FORMULATION}
The state space is represented as an object-centric scene graph. The scene graph, $\Psi$($V,E$), represents the structure of the scene. Vertices $V$ in the scene graph represent the objects present in the scene whilst the edges $E$ represent spatial relations between the objects.

We assume a perception system that takes robot observations and returns a belief $\mathcal{B}(\Psi)$ over scene graphs for the current scene. An example of such perception system is  described in Section \ref{sec:impl}. Given the task goal conditions $G$ and the current scene belief $\mathcal{B}(\Psi)$, the goal for the robot is to plan a sequence of actions $\{a_0, a_1, \dots\}$ to achieve the goal conditions $G$ under the perceptual uncertainty, and replan when needed.


\section{LESAMPLE}
The proposed planning algorithm, LESAMPLE, takes in goal conditions, $G$, and the belief over the current scene graph $\mathcal{B}(\Psi)$,  efficiently plans out a sequence of actions to achieve $G$ and replans when necessary. LESAMPLE is developed to solve problems with partially observed states and deterministic action effects.

As described in Algorithm \ref{alg:lesample}, LESAMPLE takes in goal conditions $G$ and the current belief over scene graphs $\mathcal{B}(\Psi)$ as input. LESAMPLE first samples a scene graph $\Psi_s$ from $\mathcal{B}(\Psi)$ (as described in section \ref{sec:impl}), and formulates  $\Psi_s$ and $G$ as a PDDL\cite{pddl} problem . LESAMPLE then uses a symbolic planner (Fast Downward~\cite{b10} in our implementation) to solve the PDDL problem and generate a task plan $\pi$. After taking each action in $\pi$, the robot takes a new observation $\Phi$. A validation function, $Validate(\Psi_s,a,\Phi)$, checks for inconsistency in the action effects. In particular, the validation returns $True$ if the new observation $\Phi$ after executing action $a$  matches the predicted observation based on $a$ and sampled scene graph $\Psi_s$, and returns $False$ otherwise. In our experiments, $Validate(.)$ checks if the object picked up by the robot matches the sampled scene graph hypothesis after every pick action. If the $Validate$ function returns $True$, robot continues to execute the next action in $\pi$. Otherwise, $\pi$ is discarded, the scene belief $\mathcal{B}(\Psi)$ is updated based on the observation $\Phi$ and LESAMPLE is then called recursively with the original goal condition $G$ and the updated $\mathcal{B}(\Psi)$ as parameters. LESAMPLE runs until either the last action in the current plan $\pi$ is successfully executed or timeout is reached.
  
We provide details on the continuous motion parameters used in action execution in Section \ref{sec:continuous}. 
By combining belief samples from belief space representation with the fast, goal-directed classical planning, LESAMPLE is able to efficiently plan for sequential decision making tasks with low entropy belief states.

\begin{algorithm}
\SetAlgoLined
\KwIn{Goal conditions, $G$, and Belief of current scene graph, $\mathcal{B}(\Psi)$}
\SetKwFunction{FMain}{LESAMPLE}
\SetKwProg{Fn}{Function}{:}{}
\Fn{\FMain{$G$, $\mathcal{B}(\Psi)$}}{
$\Psi_{s} ~\leftarrow ~ \text{sample from} \  \mathcal{B}(\Psi)$\\
$\pi \leftarrow ~FastDownward~ (\Psi_{s},~ G)$ \\
\ForEach{$ a \in \pi $}
        {
        $\text{take action} \ a$  \\
        $\text{take new observation} \  \Phi$\\
        $valid ~\leftarrow Validate(\Psi_{s}, a, \Phi$)\\
        \If{$valid = False$}
            {
            $\text{update belief} \  \mathcal{B}(\Psi) ~\text{given}\  \Phi$ \\
             $\text{LESAMPLE} \ (G, \mathcal{B}(\Psi))$ \\
             $\textbf{return}$
            }
        }
        \textbf{return}       
} 
\textbf{End Function}
\caption{LESAMPLE algorithm}
\label{alg:lesample}
\end{algorithm}

\section{Implementation}\label{sec:impl}
\subsection{Belief over Scene Graphs}\label{sec:belief}
For our experiments, we build a perception system that returns a $\mathcal{B}(\Psi)$ belief  over scene graphs given robot observation. We trained a Faster R-CNN~\cite{d4} detector for the 8 grocery object classes that we considered in the simulation experiments. For the $i$th detected object, the detector returns a confidence vector, which is then interpreted as the belief over object classes, $\mathcal{B}_i(c), c \in C$. $C$ is the set of all possible object classes. The spatial relations between detected objects given their detected locations are deterministically derived. With the assumption that the objects are independent from one another, we approximate the belief over scene graphs $\mathcal{B}(\Psi)$ as
$$\mathcal{B}(\Psi) \propto \displaystyle\prod_{i}^{N} \mathcal{B}_i(c)$$ where $N$ is the number of detected objects.
 
The belief over $i$th detected object, $\mathcal{B}_i(c)$, is approximated by a set of weighted object hypotheses $\{(o_k, p_k) | k=1, \dots , |C|
\}$, where $o_k =(c_k, a_k, r_k)$. For each hypothesis, $c_k$ is the object class, $a_k$ is object attributes, $r_k$ is its spatial relations with other objects in the scene graph, and $p_k$ is the weight of the hypothesis, which is equivalent to the detection confidence score corresponding to $c_k$. The object attributes $a_k$ (e.g. heavy or light) are deterministically associated with object class.

In order to draw one scene graph sample $\Psi_s$ from $\mathcal{B}(\Psi)$, we individually draw one object hypothesis $o_k$ from each $\mathcal{B}_i(c)$, such that
$$\Psi_s = \{o_k^i | i=1,\dots, N\}$$
where again $N$ is the number of detected objects. We used weighted sampling~\cite{wsample} to sample the individual object hypothesis.


\subsection{Grocery Packing Goal Conditions}
For the grocery packing task in our experiments, for each object hypotheses $o_k=(c_k, a_k, r_k)$ in the scene graph, we consider the object to be either a heavy or light, i.e. $a_k \in \{heavy, light\}$. The goal condition for the grocery packing task, which is specified as symbolic predicates in PDDL, is to have all objects packed into the box such that, heavy grocery objects are placed at the bottom of the box, and light grocery objects are placed on top of them. 

\subsection{Action schemas} \label{sec:agg}
A plan is made up of a sequence of \textit{action schemas}. An action schema consists of a set of free parameters (\texttt{\textbf{:parameters}}), conjunctive boolean pre-conditions (\texttt{\textbf{:precondition}}) that must hold for the action to be applicable  and conjunctive boolean effects (\texttt{\textbf{:effect}}) that describe the changes in the state after the action is executed. Boolean conjunctive operators used are \texttt{or, not, and}. The pick and place action schemas used in our experiments are described below:\\

\noindent \texttt{\small
 (\textbf{:action} pick \\
\noindent\textbf{:parameters} (?x)\\
\small\textbf{:precondition} (and (topfree ?x) (handempty))\\
\small\textbf{:effect}  (and (holding ?x) (not (handempty)) (not (topfree ?x)))\\
)}\\
\texttt{
(\small\textbf{:action} place\\
\small\textbf{:parameters} (?x ?y)\\
\small\textbf{:precondition} (and (holding ?x) (topfree ?y)) \\
\small\textbf{:effect} (and (not (holding ?x)) (on ?x ?y) (handempty) (not (topfree ?y)) (topfree ?x))\\
)
}

\subsection{Continuous Variables}\label{sec:continuous}
In executing actions in a plan, the robot requires certain continuous values such as collision-free arm trajectories, object grasp poses and object placement poses. We use the BiRRT \cite{birrt} motion planning algorithm to generate collision-free trajectories for \texttt{pick} and \texttt{place} actions. We determine the grasp pose of an object by querying its 6D pose from the simulator and computing a corresponding grasp configuration of the robot's gripper to pick the object from the top. The \texttt{place} actions specify destination surfaces on which to place the picked item. For example, the action \texttt{(place banana bowl)}, requires that the \texttt{banana} is placed on the surface of \texttt{bowl}. We query the Axis-Aligned Bounding Box (AABB) of both the item in hand (\texttt{banana}) and the destination surface (\texttt{bowl}) and sample legal placement poses on the surface of the AABB of the destination surface where we can place the item in hand. The first sampled legal placement pose is chosen as the placement pose of the object in hand.


\section{Quantifying Entropy}
The state space of all possible scene graphs is beyond tractability for modern computing. This space can be composed of all possible classes of objects, all possible number of objects, all possible enumerations of the 6D pose of objects,  all possible spatial relations between objects and the attributes of objects. To be computationally tractable, we make the following assumptions to constrain the state space:

\begin{itemize}
    \item The set of all possible object classes is finite and known.
    \item The perception system detects objects that are not fully occluded by other objects from the robot's field of view. Note that, the robot does not have prior knowledge of the total number of objects to expect. Thus the scene graph is made up of only detected objects and their spatial relations. 
    \item The perception system can deterministically infer the 6D pose of detected objects. Note that, there is uncertainty in the recognition of the class of detected objects.
     \item The perception system can deterministically infer spatial relations between detected objects from a 3D observation. In this work, we only consider the stacking spatial relations. In the PDDL problem description, the stacking relations are represented with axiomatic assertions \texttt{(on $o_i$ $o_j$)} for the assertion that object $o_i$ is stacked on object $o_j$ and \texttt{(topfree $o_i$)} for the axiomatic assertion that no objects are stacked on object $o_i$. 
     \item We deterministically associate detected objects with their respective attributes (either heavy or light).
\end{itemize}

As a result, the constrained state space of scene graphs, $\Psi$, will include scene graphs that have the same number of vertices as the number of detected objects, with each graph consisting of all possible enumerations of the object classes. 

We quantify the entropy of the belief over scene graphs $\mathcal{B}(\Psi)$ as the normalized sum of Shannon entropies~\cite{shannon} of the beliefs of detected objects, i.e.
\begin{equation}\label{eq:entropy}
H  =  -\frac{1}{H_{max}} \cdot \sum_{i=1}^{N}  \sum_{c \in C} p^c _i \log_{2} p^c _i
\end{equation} 
where $C$ is the set of all possible object classes, $N$ is the number of detected objects, $p^c_i$ is the probability of class $c$ of $i$th detected object in belief $\mathcal{B}_i(c)$, as explained in Section \ref{sec:belief}. $H_{max}$ is the maximum possible entropy occurring when the belief over scene graphs is uniformly distributed. i.e.
\begin{equation}\label{eq:max}
H_{max}  =  -\sum_{i=1}^{N}  \sum_{c \in C} p^c _i \log_{2} p^c _i
\end{equation} 
where $p_i ^c$ follows a uniform distribution.

For the grocery packing task we consider in this work, we use Equation \ref{eq:entropy} to classify the perceptual entropy levels of grocery packing tasks.  On one extreme, $H = 0$ for a scene graph estimate with no uncertainty. On the other extreme, $H = 1$ for a scene graph with high uncertainty. 
In our experiments, we perform grocery packing on scene graphs with $H$ values from $0.1$ to $0.9$. 


\section{EXPERIMENTS}
\begin{figure}[t!]
    \centering
    \scalebox{0.6}{\includegraphics[trim=4 4 4 4,clip]{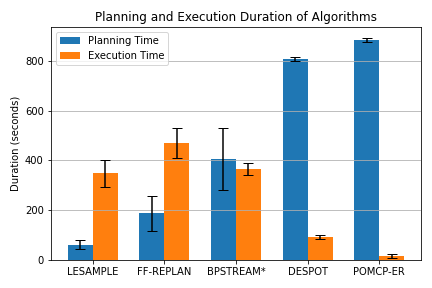}}
    \caption{\textbf{Planning and Execution time} results for  LESAMPLE and the benchmarked algorithms from performing the low entropy Grocery Packing Tasks (\textit{H} values between 0.3 and 0.5). Error bars represent one standard deviation from the mean. The maximum time allocated for each task is 900 seconds.}
    \label{fig:timing}
\end{figure}

\begin{figure}[t!]
    \centering
    \scalebox{0.5}{\includegraphics[trim=4 4 4 4,clip]{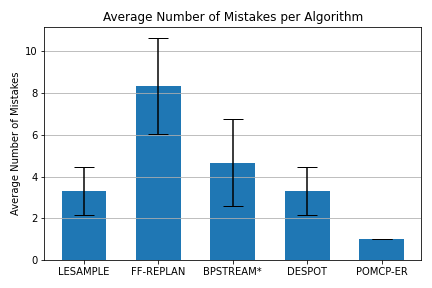}}
    \caption{Experimental results for \textbf{Number of Mistakes} for  LESAMPLE and the benchmarked algorithms from performing the low entropy Grocery Packing Tasks (\textit{H} values between 0.3 and 0.5). Error bars represent one standard deviation from the mean. Note that POMCP-ER and DESPOT make few mistakes because the planning time takes up most of the allocated time per task. Hence they barely take any actions before timeout is reached.}
    \label{fig:mistakes}
\end{figure}

\begin{figure}[t!]
    \centering
    \scalebox{0.5}{\includegraphics[trim=4 4 4 4,clip]{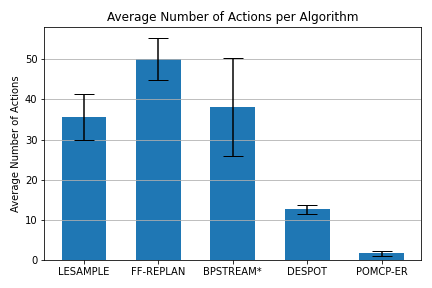}}
    \caption{ Experimental results for \textbf{Number of Actions} for  LESAMPLE and the benchmarked algorithms from performing the low entropy Grocery Packing Tasks (\textit{H} values between 0.3 and 0.5). }
    \label{fig:actions}
\end{figure}

\begin{figure}[t!]
    \centering
    \scalebox{0.5}{\includegraphics[]{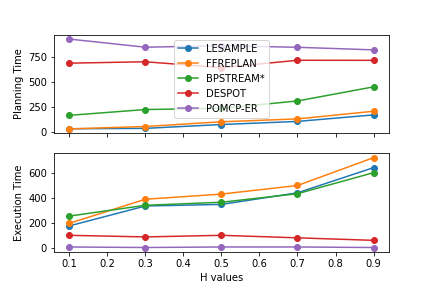}}
    \caption{ Experimental results for the \textbf{Planning Times and Execution Times} for  LESAMPLE and the benchmarked algorithms from performing the Grocery Packing Task for increasing entropy values (\textit{H} values from 0.1 to 0.9). The maximum time allocated for each task is 900 seconds. DESPOT and POMCP-ER do not complete any of the tasks before the 900 second time-out for any of the \textit{H} values.}
    \label{fig:entropies}
\end{figure}

We compare the performance of LESAMPLE with replanning and belief space planning methods on low entropy grocery packing tasks ($H$ values between $0.3$ and $0.5$), shown in results in Table \ref{table:results} and  Figures \ref{fig:timing}, \ref{fig:mistakes} and \ref{fig:actions} and a broader range of entropy values ($H$ values from $0.1$ to $0.9$), shown in results in Figure \ref{fig:entropies}. The simulation environment we use is depicted in Figure \ref{fig:1}. 
The goal condition for the grocery packing task is to have all items packed into the box such that, heavy grocery items are placed at the bottom of the box, and light grocery items are placed on top of them.
Experiments were run in the Pybullet simulation~\cite{d1}. We use a simulated Digit robot \cite{digit} equipped with suction grippers on both arms. 8 3D models of grocery items from the YCB dataset~\cite{d2} were used as grocery items in the experiments. A Faster R-CNN~\cite{d4} object detector is trained to detect these grocery items and return a confidence score vector for each detected object. We normalize these confidence scores, add entropy based on the specific \textit{H} value of the task, as prescribed in Equation \ref{eq:entropy}, and form the belief over scene graphs $\mathcal{B}(\Psi)$. The experiments were run on a laptop with 2.21GHz Intel Core i7 CPU, 32GB RAM and a GTX 1070 GPU. We also demonstrate our approach on a real world grocery packing task with a physical Digit robot. A summary of our approach as well as the real world demo can be seen in the accompanying video and at this url: {\url{https://youtu.be/im6tve9-9A0}}.

The following methods are benchmarked in this experiment:
\begin{itemize}

    \item \textbf{FF-REPLAN}: This algorithm \cite{b6} performs symbolic planning on a determinized belief over detected objects. It determinizes the belief over scene graphs by choosing the hypothesis with the highest probability in $\mathcal{B}_i(c)$ for each detected object, $i$. The algorithm then formulates the determinized scene graph and the goal conditions (Section V-B) as a PDDL~\cite{pddl} problem and solves it using Fast Downward~\cite{b10} to generate a plan to pack objects into the box. The algorithm uses the \textit{Validate} function (same as in LESAMPLE in Algorithm \ref{alg:lesample}) to check if the object picked up by the robot matches the determinized scene graph after every pick action. If \textit{Validate} returns \textit{True}, the next action in the plan is executed. If \textit{Validate} returns \textit{False}, a replan request is triggered. The belief over detected objects is updated and determinized again. A new plan is generated accordingly. The robot plans and executes until either all the objects are packed or the 15-minute timeout is reached.
    
    \item \textbf{LESAMPLE}: This algorithm performs LESAMPLE on the belief over detected objects to pack them into the box. LESAMPLE terminates either when all the objects are packed into the box or when the 15-minute timeout is reached.
    
    \item \textbf{BPSTREAM*}: This algorithm is a variation of BeliefPDDLStream \cite{beliefpddlstream} adapted to suit our grocery packing task. We replace the streams in the original BeliefPDDLStream with simple parameter samplers for sampling motion plans and other continuous action parameters as described in Section \ref{sec:continuous}. We use a set of 8 weighted particles to represent the belief of each detected object.
    
    \item \textbf{POMCP-ER}: POMCP-ER is a variation of the POMCP \cite{pomcp} belief space planning algorithm  with episodic rewards. Here, the robot receives a reward of 10 whenever an item is packed into the container, a reward of 100 when the arrangement satisfies the packing conditions and a reward of -10 when the arrangement fails to satisfy the packing conditions. POMCP-ER is also restricted to 10 iterations, each with a rollout depth of 10 and represents the belief set with 10 particles. Since grocery packing is a goal-directed task, we set the discount factor to 1, thus future rewards are just as valuable as immediate ones. To narrow the focus of the Monte Carlo Tree Search in POMCP, as prescribed by~\cite{pomcp}, we use domain knowledge by specifying the subset of preferred actions at each node in the search tree.
    
    \item \textbf{DESPOT}: We use an anytime and regularized version of the DESPOT belief space planning algorithm\cite{despot}. DESPOT improves upon POMCP's poor worst case behavior by  sampling a small number of scenarios (3 scenarios in our implementation) and searching over a determinized sparse partially observable tree. Here, we use the same reward function, maximum number of iterations, maximum rollout depth and discount factor as POMCP-ER.  
\end{itemize}

\begin{figure}[t!]%
    \centering
     \subfloat[]{{\includegraphics[width=0.15\textwidth]{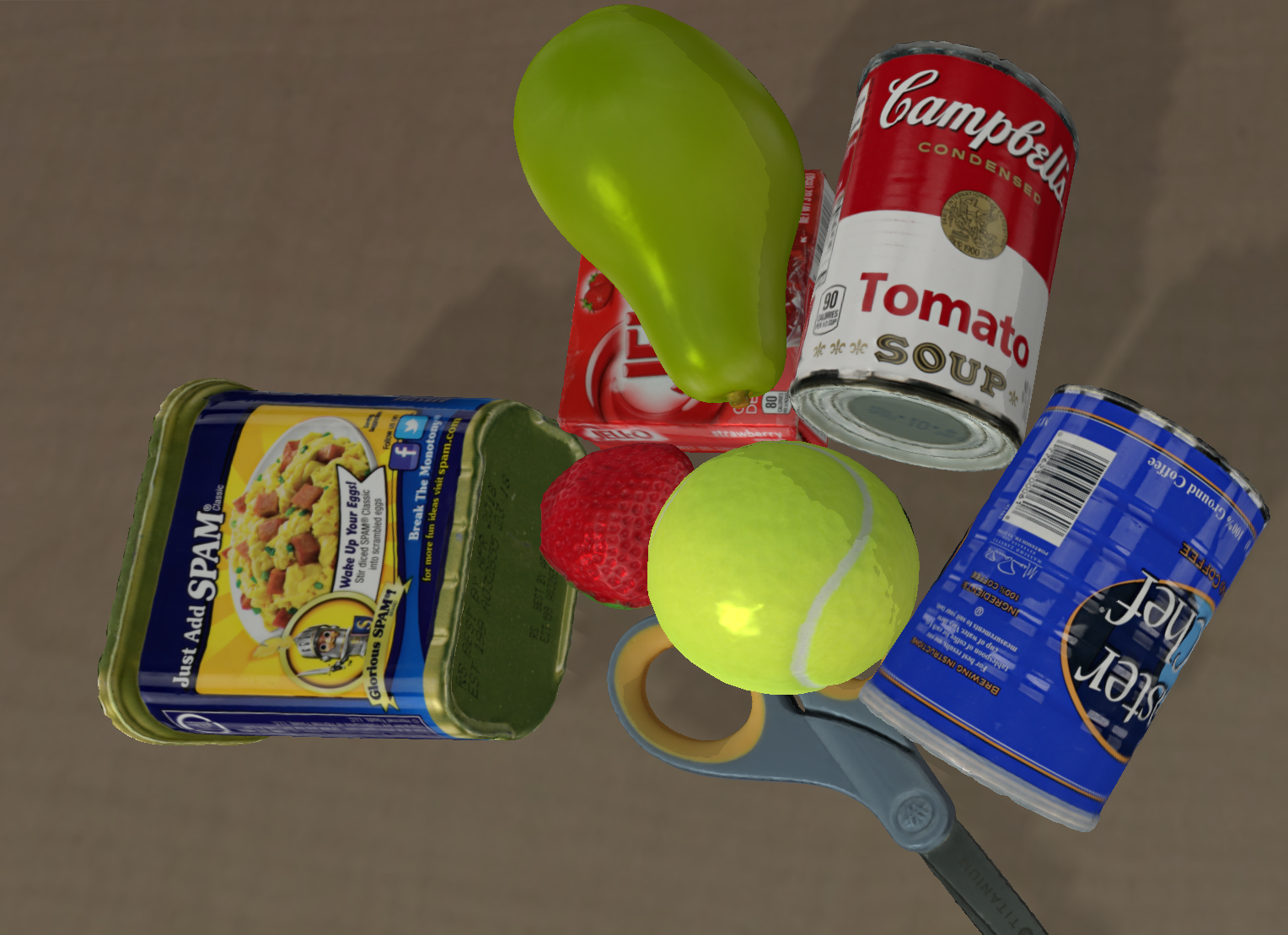} }}%
    \qquad
     \subfloat[]{{\includegraphics[width=0.15\textwidth]{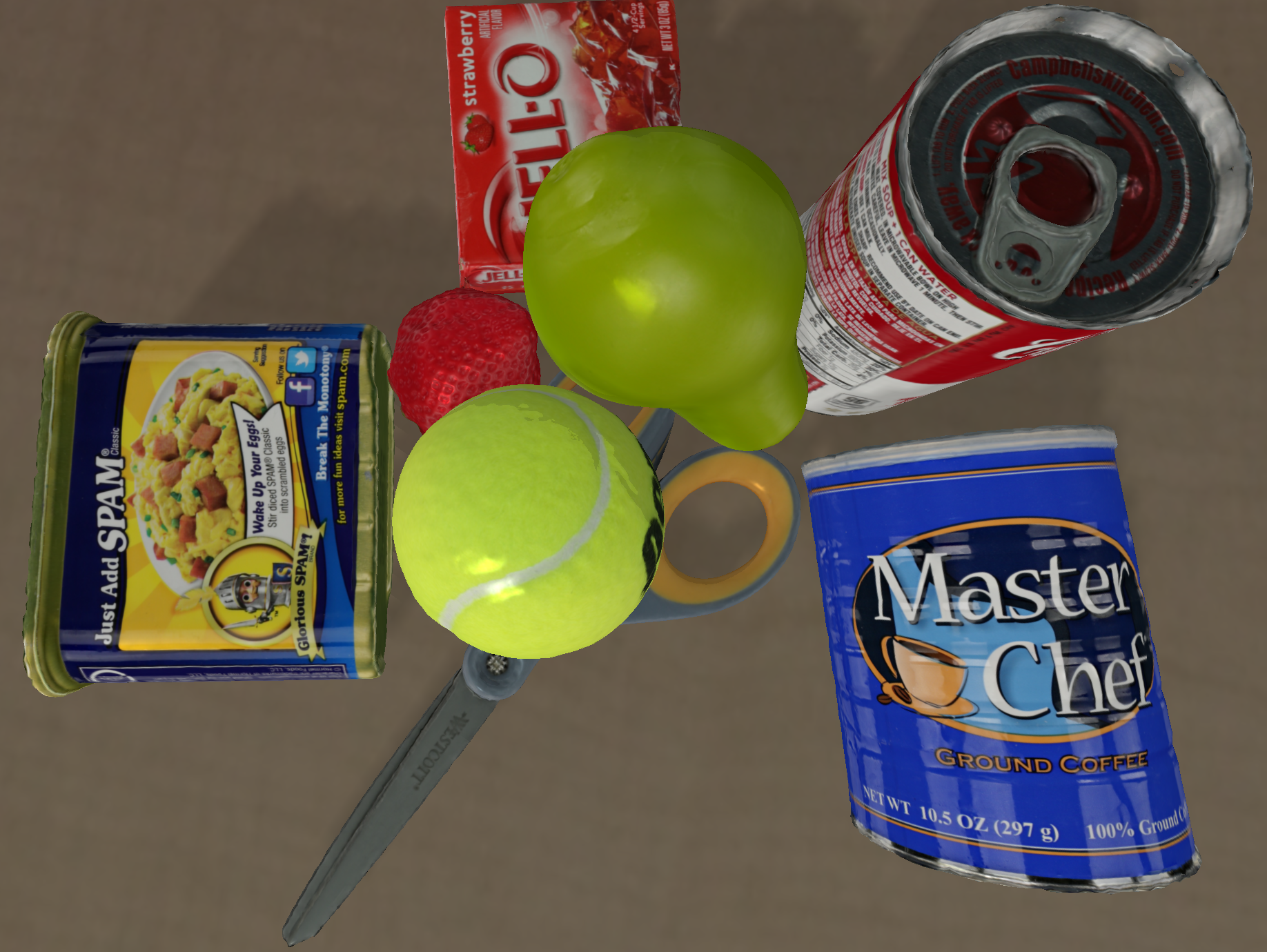} }}%
    \caption{ Examples of initial cluttered scenes}%
    \label{fig:arr}%
\end{figure}

The methods are benchmarked on low-entropy Grocery Packing tasks. Their performance results are displayed in Table \ref{table:results} and  Figures \ref{fig:timing}, \ref{fig:mistakes} and \ref{fig:actions}. 

We run each planning method on 5 different initial arrangements of the grocery items, examples of which are showin in Figure \ref{fig:arr}. In Figure \ref{fig:timing}, we show planning time and execution time averaged across 5 initial scenes. For each initial scene, we run each method 5 times.  

As shown in Table \ref{table:results} and Figures \ref{fig:timing}, \ref{fig:mistakes} and \ref{fig:actions}, across all tasks in our experiments, LESAMPLE  outperforms other baseline methods by having the least completion times, making the least number of mistakes and as such requiring the least number of pick-and-place actions to complete the task. BPSTREAM* has a slightly higher execution time than LESAMPLE and a significantly higher planning time than both FF-REPLAN and LESAMPLE. The belief space planning algorithms, DESPOT and POMCP-ER, have the worst results for every metric. DESPOT and POMCP-ER make fewer number of mistakes because they spend majority of the time planning and are only able to take a few actions before the 15 minute timeout is reached. DESPOT however performs slightly better than POMCP-ER and is able to successfully pack over half of the groceries before the 15 minute timeout.

FF-REPLAN chooses the most likely hypothesis and disregards the inherent entropy in the state space. As a result, FF-REPLAN makes a mistake when the most likely hypothesis does not correspond to the true state. On the other hand by employing the belief space representation of belief space planning, LESAMPLE is able to maintain a belief of the various scene hypotheses and update this belief in the next replanning cycle even after it samples a false hypothesis. This makes LESAMPLE more robust to noisy state estimation. 
It is worth noting that, in scenarios where the most likely hypothesis of the scene estimate represents the true scene graph, FF-REPLAN performs less number of actions than LESAMPLE. This is because LESAMPLE does not always sample the true scene graph hypothesis and could potentially perform more actions than necessary. Such scenarios however do not occur often enough in the low entropy tasks to make FF-REPLAN a more efficient approach than LESAMPLE.

BPSTREAM* aggressively performs online planning. It only executes the first action in the generated plan and replans even when no mistake is committed. As such, even thoug
h BPSTREAM* employs the belief space representation and samples from the belief space, it ends up planning much more often LESAMPLE, resulting in its higher planning time. Because it samples from the belief space, BPSTREAM* commits less mistakes than FF-REPLAN.

\begin{table}
\centering
\resizebox{\linewidth}{!}{
\begin{tabular}{|c||c|c|c|}
\hline 
\textbf{Algorithm} & Total time spent(s) & Avg. time per action & Avg. num. packed items  \\
\hline \hline
LESAMPLE &  \textbf{411.3} & \textbf{11.5} & \textbf{8.0} \\ 
  \hline
FF-Replan & 658.3 & 13.2 & 8.0 \\
\hline
BPSTREAM* & 770.4 & 20.3 & 8.0 \\  
\hline
DESPOT & 900.0**  & 71.0 & 4.5   \\
\hline
POMCP-ER & 900.0** & 539.9 & 1.0 \\ 
\hline
\end{tabular}}
\caption{Summary of results from experiments for low entropy tasks (\textit{H} values between 0.3 and 0.5). **DESPOT and POMCP-ER could not complete the tasks before the 900 second timeout so we set their total time spent to 900 seconds}
\label{table:results}
\end{table}

POMCP-ER performs the worst in planning time, execution time and number of items packed. It is unable to complete any of the packing tasks. It also packs significantly fewer objects than LESAMPLE, FF-REPLAN and BPSTREAM*. This is because of the high branching factor in the Monte Carlo search tree.  DESPOT searches over a sparser tree than POMCP-ER, making it faster and better performing than POMCP-ER. DESPOT however still falls short when compared with BPSTREAM*, FF-REPLAN and LESAMPLE which perform symbolic planning on a determinized belief space.

Belief Space Planning approaches like POMCP-ER and DESPOT are developed to solve problems with high entropy state spaces. As such, they spend a lot of computation in deciding the approximately optimal action to take next. This makes them inefficient for state spaces with low entropy. By adopting the fast, goal-directed features of classical planning through the use of a symbolic planner, LESAMPLE is able to efficiently solve low entropy problems by directly acting upon a sampled scene graph. This makes LESAMPLE a favorable choice for solving low perceptual entropy tasks.  

We also compare the average planning and execution times of LESAMPLE with the benchmarked algorithms for tasks with increasing perceptual entropy as shown in the results in Figure \ref{fig:entropies}. FF-REPLAN has the least planning time at the lowest \textit{H} value because at such entropy levels, FF-REPLAN makes little to no mistakes and as such, does not have to re-plan often. As the \textit{H} value increases however FF-REPLAN expends more planning and execution time than LESAMPLE because it makes more mistakes and replans more often. BPSTREAM* consistently spends more planning time than LESAMPLE across all \textit{H} values because it, by design, replans after every action, whether or not a mistake is committed. DESPOT and POMCP-ER do not complete any of the tasks across the \textit{H} values before the 900 second time-out, resulting in their relatively lower execution times. Even though DESPOT and POMCP-ER are belief space planning approaches designed for high entropy state spaces, they still spend significant amounts of time to plan for a single action and are, as a result, out-performed by fast replanning approaches even in high entropy scenarios in our experiments. As such, for reversible tasks like grocery packing where action effects could be reversed through taking other actions, it is more efficient in the long run to employ fast replanning approaches that are able to quickly plan actions and replan to recover from mistakes. This resonates with results by Little et. al. \cite{b7}. However, for tasks where action effects are irreversible or where the penalty for making mistakes is significant, the more deliberative belief space planning approaches like DESPOT and POMCP-ER are more appropriate.

\section{CONCLUSION}

We presented LESAMPLE as an online planning method for efficiently solving sequential decision-making tasks with low perceptual entropy. The key idea is to use classical planning on estimates resulting from belief space inference over perceptual observations.  As a result, LESAMPLE can perform more efficient goal-directed reasoning under scenarios of low-entropy perception. We demonstrated the efficiency of this method on grocery packing tasks. LESAMPLE demonstrated advantages in low-entropy scenarios where classical planning cannot handle uncertainty and belief space planning is unnecessarily computationally expensive.


%

\ifCLASSOPTIONcaptionsoff
  \newpage
\fi



%

%

\end{document}